\documentclass[conference]{IEEEtran}

\makeatletter
\def\endthebibliography{%
	\def\@noitemerr{\@latex@warning{Empty `thebibliography' environment}}%
	\endlist
}
\makeatother
\IEEEoverridecommandlockouts
\usepackage{cite}
\usepackage{amsmath,amssymb,amsfonts}
\usepackage{algorithmic}
\usepackage{graphicx}
\usepackage{textcomp}
\usepackage{xcolor}
\usepackage{comment}
\usepackage{booktabs}       
\def\BibTeX{{\rm B\kern-.05em{\sc i\kern-.025em b}\kern-.08em
    T\kern-.1667em\lower.7ex\hbox{E}\kern-.125emX}}
\begin{document}

\title{A Biologically Plausible Audio-Visual Integration Model for Continual Learning
}

\author{\IEEEauthorblockN{Wenjie Chen$^{1,2}$, Fengtong Du$^2$, Ye Wang$^{1,2}$, Lihong Cao$^{1,2,3,*}${\footnotesize }
		\thanks{{*}Corresponding author: lihong.cao@cuc.edu.cn}
	}	
\IEEEauthorblockA{
$^1$\textit{State Key Laboratory of Media Convergence and Communication, Communication University of China, Beijing, China}\\
$^2$\textit{Neuroscience and Intelligent Media Institute, Communication University of China, Beijing, China}\\
$^3$\textit{State Key Laboratory of Mathematical Engineering and Advanced Computing, Wuxi, China}\\
}
\{jessiechen, dft, yewang, lihong.cao\}@cuc.edu.cn\\
}

\maketitle

\begin{abstract}
The problem of catastrophic forgetting has a history of more than 30 years and has not been completely solved yet. Since the human brain has natural ability to perform continual lifelong learning, learning from the brain may provide solutions to this problem.
In this paper, we propose a novel biologically plausible audio-visual integration model (AVIM) based on the assumption that the integration of audio and visual perceptual information in the medial temporal lobe during learning is crucial to form concepts and make continual learning possible. 
Specifically, we use multi-compartment Hodgkin-Huxley neurons to build the model and adopt the calcium-based synaptic tagging and capture as the model's learning rule.
Furthermore, we define a new continual learning paradigm to simulate the possible continual learning process in the human brain. We then test our model under this new paradigm. Our experimental results show that the proposed AVIM can achieve state-of-the-art continual learning performance compared with other advanced methods such as OWM, iCaRL and GEM. Moreover, it can generate stable representations of objects during learning. These results support our assumption that concept formation is essential for continuous lifelong learning and suggest the proposed AVIM is a possible concept formation mechanism. 

\end{abstract}

\section{Introduction}
The problem of catastrophic forgetting has a history of more than 30 years\cite{McCloskeyCF}, and has always been a barrier to the development of lifelong learning artifical intelligence.

The past few years have seen the rapid development of deep learning and the community of continual learning. Under this background, a series of continual learning paradigms (such as incremental learning of new instances, new classes, and the mixture of both) have emerged \cite{lomonaco2017core50:} and a variety of methods have been proposed by different research groups around the world. The mainstream methods proposed for continual learning are within the framework of artificial neural network (ANN) with backpropagation (BP) algorithm \cite{rumelhart1988learning}. Broadly, these methods can be organized into the following topics: regularizations of the network \cite{goodfellow2013an-r,kirkpatrick2017overcoming-r,zenke2017continual-r,lee2017overcoming-r,zeng2018continuous-r,Li2018LwF-r,golkar2019continual-r}, parameters isolation \cite{mallya2018packnet-i:}\cite{fernando2017pathnet-i:}, dynamic structure \cite{charles2017progressive-d}\cite{aljundi2017expert-d}, memory-based consolidation \cite{lopezpaz2017gradient-m}\cite{rebuffi2017icarl:-m}, and the dual memory systems inspired by the complementary learning system theory in the brain \cite{mcclelland1995why-CLS,mcclelland2020integration-CLS,kemker2018fearnet-CLS}. Although many new methods has been proposed, they are still far from the level of human continual learning. How to achieve human-level continual lifelong learning remains to be explored.

\textbf{Learning from the brain.}
The causes of catastrophic forgetting problem can be manifold, so the solution to this problem needs to be considered in many ways.

From the perspective of the neuron model, McCulloch-Pitts neuron in ANN is oversimplified compared with the neurons in the brain \cite{hodgkin1952a}. Neurons in the brain are not just points, they have tree-like dendrites that perform nonlinear computations \cite{major2013active}. The dendrites of pyramidal cells in the human brain are more complex than those of other species \cite{beaulieularoche2018enhanced}, suggesting that the multi-compartmental neuron model might be an essential structure of high-level intelligence. 

From the perspective of the synaptic model, the synapses in ANN are also oversimplified compared to the real synapses in the brain. As for the excitatory and inhibitory synapses in the brain, their functions are not merely doing addition or subtraction. Spatial and temporal correlations should also be concerned when integrating the neural signals at the synapses.

Last, but perhaps most important, is synaptic plasticity. Although BP is very efficient in ANN, it is still controversial that whether BP is biologically plausible in the brain \cite{whittington2019theories}\cite{Timothy2020BP}. Synaptic plasticity in the brain involves the proteins synthesis in the postsynaptic neurons, and is closely related to the change of calcium concentration \cite{rogerson2014synaptic}. The theory of synaptic tagging and capture (STC) depicts the relationship between calcium concentration and plasticity-related proteins. It provides an elegant biological explanation of the consolidation of newly formed memories at the cellular and synaptic scale \cite{redondo2011making}. The computational model \cite{clopath2008tag-trigger-consolidation:}\cite{smolen2012molecular} of STC can also explain some experimental phenomena well. 
\begin{figure*}	[t]
	\vspace{-0.3cm}
	\centering
	\includegraphics[width=0.72\textwidth]{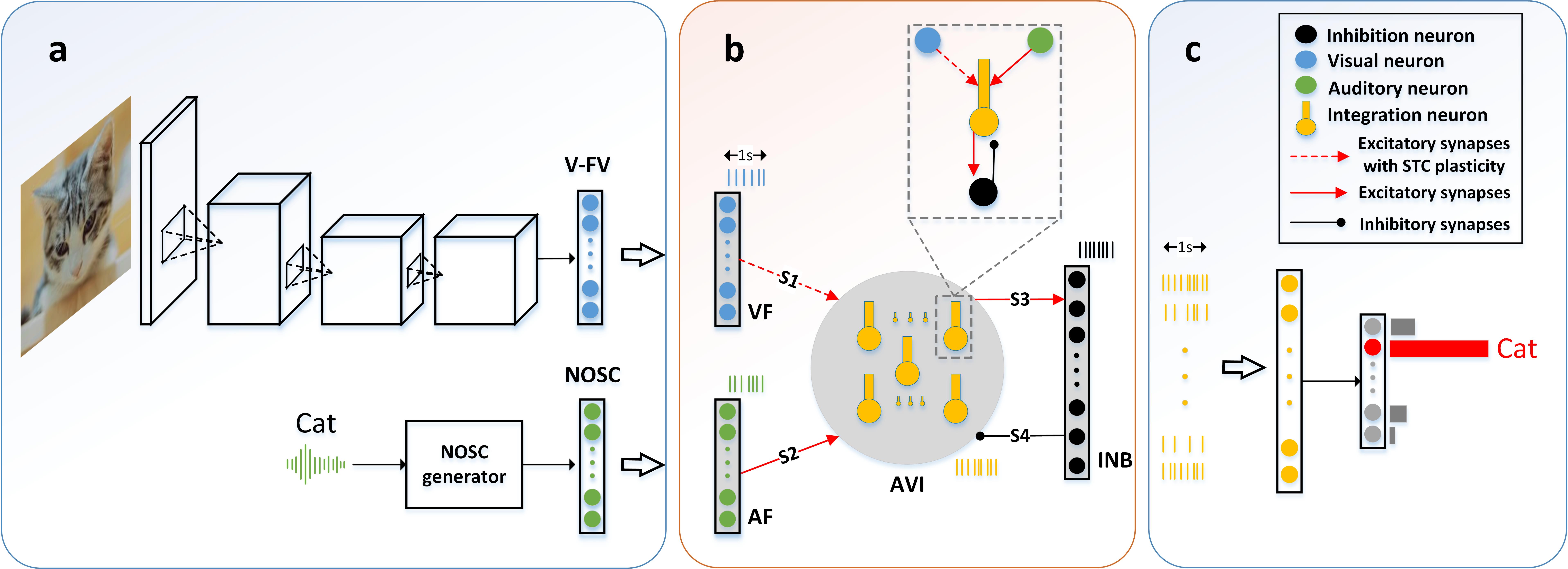}
	\caption{The overall framework of the present work for continual learning. \textbf{(a)}. Encoding module. \textbf{(b)}. Integration module. \textbf{(c)}. Decoding module.}
	\label{overallArchitecture}
	\vspace{-0.7cm}
\end{figure*}

\textbf{Concept formation and continual learning}.
For humans, the result of continual learning in the brain is the formation of concepts. Concept cells in the human brain are first discovered and discussed in the literature \cite{quiroga2012concept}. These amazing cells, which are discovered in the medial temporal lobe (MTL) in the brain, respond to multi-modal stimuli of specific concepts. For instance, Jennifer Aniston's cells respond to any stimuli closely related to Jennifer Aniston, whatever modality the stimuli belong to, and would be silence if they encounter other stimuli. Moreover, according to \cite{ForgettingMaching}, these concept cells seem to be unique in the human brain and may not exist in the brains of other animals due to the lack of human-like high-level language functions. Considering that humans always have better continual lifelong learning ability than other animals, we assume that these concept cells in the human brain may play an important role to make continual lifelong learning possible. Although the concept formation mechanism is still unknown, exploring its possible mechanism is vital in comprehending the meaning of specific concepts in the human brain. Based on the assumption above, from the perspective of building up a human-like continual learning system, we find it necessary to establish the relationship between continual learning and concept formation, and model the possible process of concept formation in the brain under a continual learning paradigm. 

In order to do the simulation, we focus on the multi-modal integration in the perirhinal cortex of the MTL. The perirhinal cortex is a primary source to the entorhinal cortex, which is the main interface between the hippocampus and neocortex. Moreover, the cells in this region also respond to multi-modal stimuli \cite{taylor2006binding}\cite{clarke2014object}, which might be the origin of concept cells in the hippocampus. In terms of multi-modal inputs, the perirhinal cortex receives visual signals mainly from the temporal area TE, which is the end of the ventral visual pathway, and it receives auditory signals mainly from the parahippocampus, which has strong connections with the auditory cortex \cite{suzuki2014the}.

\textbf{Method overview.} Inspired by the multi-modal integration in the perirhinal cortex of the MTL, we propose a novel biologically plausible audio-visual integration model (AVIM) for modeling concept formation in a continual learning paradigm. Besides, we design a set of pre-defined high-level auditory feature codes to guide the integration learning based on the following three assumptions: (i) The auditory system of the brain is capable of self-learning \cite{vallabha2007unsupervised}; (ii) The development of hearing is earlier than vision; (iii) Hearing has a significant impact on vision. Under this circumstance, the whole integration learning process can be regarded as a supervised continual learning process, with the supervised signals coming from the pre-defined auditory feature codes.

Fig. \ref{overallArchitecture} shows the proposed overall framework for continual learning. In general, there are three modules in our model: the encoding module, the integration module, and the decoding module. The encoding module's role is to preprocess the sensory signals, including extracting the high-level features of visual signals and generating a set of pre-defined auditory feature codes served as the supervised signals for learning. The integration module's role is to do audio-visual associative learning and generate audio-visual concept representations under a continual learning paradigm. Finally, the outputs of the integration layer are fed to the decoding module to obtain final classification results. 

The contributions of the present work can be summarized as the following: 1. We propose a biologically plausible audio-visual integration model as the solution to the continual learning task defined in Section \ref{DefineCLParadigm}. 2. We define NOSC as the high-level auditory object feature codes and use it to guide the audio-visual integration in the continual learning paradigm; 3. We design an energy normalized linear classifier to decode spike trains from the audio-visual integration layer and an updating algorithm for this classifier to overcome catastrophic forgetting in the continual learning process. 

The organizational structure of the rest paper is as follows. In Section \ref{Method}, we will explain the details of the overall framework for concept formation and continual learning. In Section \ref{ExperimentalDesign} and Section \ref{ImplementationDetail}, we will describe the experimental design and the implementation details, respectively. In Section \ref{Results} and \ref{Discussion}, we will analyse the experimental results of our model and other advanced methods and discuss the advantages and possible limitations of the proposed model. Finally, we will draw a conclusion in Section \ref{Conclusion}.
\section{Method}
\label{Method}
In this section, we will describe the three modules in the proposed model, namely the encoding module, the integration module and the decoding module in order.
\subsection{Encoding module}
The role of the encoding module is to extract the high-level features of sensory signals. Specifically, for visual signals, we use a pre-trained convolutional neural network (CNN) to get high-level visual features, denoted as V-FV. For a given CNN and an input image $I$, we define the V-FV of $I$ to be the output vector of the last layer before classification layer of this CNN. The length of a V-FV is denoted as $N_{vfv}$.
\begin{figure}[h]	
	\centering
	\includegraphics[width=0.2\textwidth]{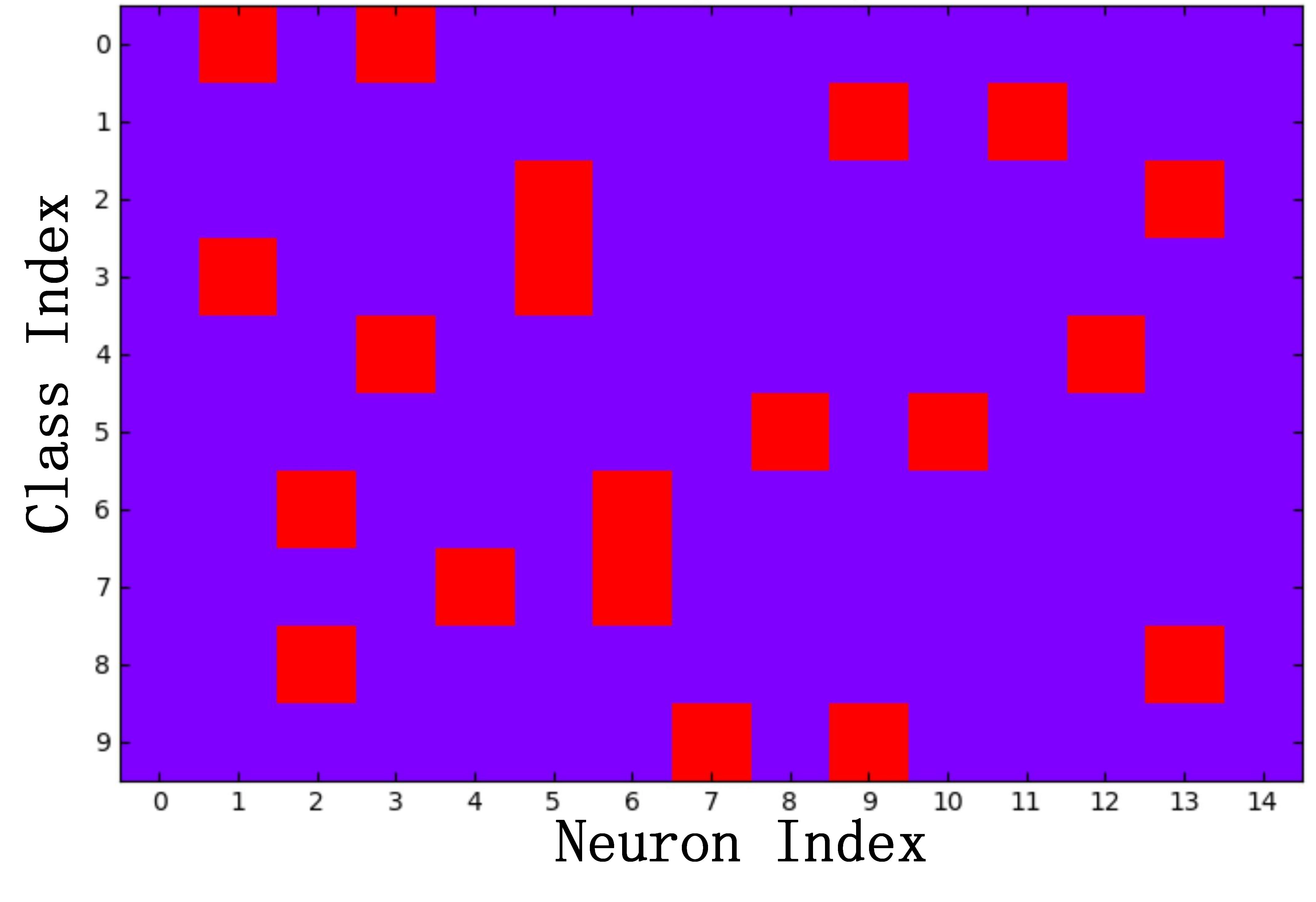}
	\vspace{-0.3cm}	
	\caption{Visualization of the NOSC(10, 15, 2, 1) under a specific random seed. Red squares represent neurons with value of 1. }
	\label{nosc10_15_2_1}
	\vspace{-0.6cm}	
\end{figure}

For auditory signals, we define the high-level auditory feature codes of different concepts to be a set of near-orthogonal and sparse binary codes, denoted as NOSC for simplicity. We assume that these auditory signals serve as a kind of supervised signal for vision during learning. Several facts support the NOSC assumption. Firstly, high-level representations of auditory objects in the brain are highly sparse coded. Secondly, the number of shared neurons in the high-level representations of auditory objects are relatively small \cite{waydo2006sparse}. A set of NOSCs for $C$ concepts can be written as \emph{NOSC(C, $N_{nosc}$, n, K)}, where \emph{$N_{nosc}$} is the length of a NOSC, \emph{n} is the number of neurons with value of one in a NOSC and \emph{K} is the max number of shared neurons between any two NOSCs. See Fig \ref{nosc10_15_2_1} for an example of \emph{NOSC}(10, 15, 2, 1). 

\subsection{Integration module}
The role of the integration module is to do audio-visual integration under a continual learning paradigm, as shown in Fig. \ref{overallArchitecture} (b). We name this module AVIM (audio-visual integration model) for simplicity. There are four layers in AVIM: the VF, AF, AVI, and INB. Here, VF, AF, AVI and INB represent the visual feature layer, the auditory feature layer, the audio-visual integration layer, and the inhibition layer, respectively. Amongst these four layers, AVI is the most critical one, it comprises a set of multi-compartment Hodgkin-Huxley neurons, whose dendrites receive high-level visual and auditory signals from VF and AF, respectively. Besides, we design INB connecting to AVI for balancing its energy. 
\paragraph{Network structure}
For the sake of description, we denote the number of neurons in VF, AF, AVI and INB as $N_{v}$, $N_{a}$, $N_{av}$, and $N_{i}$, respectively.  Furthermore, we denote the connections from VF to AVI as $S_{1}$, the connections from AF to AVI as $S_{2}$, the connections from AVI to INB as $S_{3}$ and the connections from INB to AVI as $S_{4}$. 

In AVIM, the proportion of neurons between different layers have the following relations: $N_{v}$ = $N_{a}$, $N_{av}$ = 3$N_{v}$ and $N_{i}$ = 0.25$N_{av}$. The projecting ratio of $S_{1}$, $S_{2}$, $S_{3}$, and $S_{4}$ are set to be 1:4, 1:6, 1:1.5, and 1:10, respectively. For example, if the projecting ratio of $S_{1}$ is set to 1:4, it means that each neuron in VF is connected to four neurons in AVI on average. 
\paragraph{Neuron model}
For neurons in VF and AF, we use point neuron model. 
Specifically, the firing rate of the \emph{j-th} neuron in VF/AF is 20$v$, here $v$ ($v$$\in$[0,1]) is the activation value of the corresponding \emph{j-th} neuron in V-FV/NOSC generated from the encoding module. In addition, the spike trains of neurons in VF/AF follow the Poisson distribution.

For neurons in AVI, we adopt a two-compartment pyramidal neuron model proposed in \cite{yuen1991reconstruction} based on the fact that the major neurons in the perirhinal cortex are pyramidal cells. The neuron in AVI consists of a soma and a dendrite. The dynamic equations of voltage in soma and dendrite are described in \eqref{SomaDV} and \eqref{SynDV}, respectively. The constant parameters in \eqref{SomaDV} and \eqref{SynDV} are: ${C}_{m}=3.4\mu F/{cm}^{2}$, ${g}_{ds}=0.11\mu S/{cm}^{2}$, ${g}_{sd}=0.33\mu S/{cm}^{2}$. More details of these dynamic equations can be seen in  \cite{yuen1991reconstruction}.

For neurons in INB, we adopt an inhibitory neuron model in the hippocampus proposed in \cite{wang1996gamma}, which has one compartment, see \eqref{INBSomaDV}. The constant parameter in \eqref{INBSomaDV} is ${C}_{m}=1\mu F/{cm}^{2}$.
\begin{equation}
\begin{split}	
{C}_{m}\frac{d{V}_{s}}{dt}=&-{I}_{L}-{I}_{Na}-{I}_{K}-{I}_{Ca}-{I}_{ahp}\\
&-{g}_{ds}({V}_{s}-{V}_{d})+{I}_{synToSoma}
\label{SomaDV}
\end{split}
\end{equation}
\begin{equation}
\begin{split}
{C}_{m}\frac{d{V}_{d}}{dt}=&-{I}_{L}-{I}_{Na}-{I}_{K}-{I}_{Ca}-{I}_{ahp}\\
&-{g}_{sd}({V}_{d}-{V}_{s})+{I}_{synToSoma}
\end{split}
\label{SynDV}
\end{equation}
\begin{equation}
{C}_{m}\frac{dV}{dt}=-{I}_{L}-{I}_{Na}-{I}_{K}+{I}_{syn}
\label{INBSomaDV}
\end{equation}
\paragraph{Synaptic model}
As for connections in AVIM, $S_{1}$ is the excitatory connections with AMPA and NMDA receptors. $S_{2}$ and $S_{3}$ are excitatory connections with only AMPA receptors. $S_{4}$ is inhibitory connections with GABA receptors. AMPA and GABA receptors are ligand-gated ion channels, and their synaptic current models follow \eqref{ligand-gated}. Here, $g_{syn}$ is the receptor conductance, ${E}_{syn}$ is the reverse potential of the receptor.
\begin{equation}
{I}_{syn}(t)={g}_{syn}(t)({V}_{m}(t)-{E}_{syn})
\label{ligand-gated}
\end{equation}
NMDA receptor is a voltage-gated ion channel, and its synaptic current model is shown as \eqref{voltage-gated}. Here, $[{Mg}^{2+}]_{o}=1$, $\beta=0.08$ and $\gamma=9$.
\begin{equation}
\begin{split}
{I}_{syn}(t)&={g}_{syn}(t)s(V)({V}_{m}(t)-{E}_{syn})\\
s(V)&=\frac{1}{1+[{Mg}^{2+}]_{o}exp(-\beta{V}_{m}+\gamma)}
\end{split}
\label{voltage-gated}
\end{equation}
The model of receptor conductance $g_{syn}$ is a $\beta$-function, see the following \eqref{betaFunction}. Here, $\overline{g}_{syn}$ is the maximum receptor conductance,  ${\tau}_{rise}$ and ${\tau}_{decay}$ are the time constants, and $x(t)$ are the spike trains of presynaptic neuron.
\begin{equation}
{\tau}_{rise}{\tau}_{decay}\frac{d^2g}{dt^2}+({\tau}_{rise}+{\tau}_{decay})\frac{dg}{dt}+g=\overline{g}_{syn}x(t)
\label{betaFunction}
\end{equation}
Besides, the reversal potentials of the above receptors are: ${E}_{AMPA}=0mV,{E}_{NMDA}=0mV,{E}_{GABA}=-80mV$.
See TABLE \ref{synapticModelSettings} for the details of parameters settings of the synaptic models.
\vspace{-0.4cm}
\begin{table}[h]
	\caption{Parameters settings of the synaptic models. }
	\centering
	\vspace{-0.2cm}
	\resizebox{88mm}{10mm}
	{
		\begin{tabular}{cccccc}
			\toprule
			\emph{Synapses}     &\emph{$\overline{g}_{AMPA}(mS/cm^2)$}	&\emph{$\overline{g}_{NMDA}(mS/cm^2)$}	&\emph{$\overline{g}_{GABA}(mS/cm^2)$}	&\emph{${\tau}_{rise}(ms)$} &\emph{${\tau}_{decay}(ms)$}\\
			\midrule
			$S_{1}$ &0.1	&0.1	&N	&5   &100 \\
			$S_{2}$ &1.0	&N	    &N	&2    &2	\\
			$S_{3}$ &0.01 &N	    &N	&2    &2	\\
			$S_{4}$ &N  &N      &0.0002 &5 &100 \\
			\bottomrule
			\label{synapticModelSettings}
		\end{tabular}
	}
\end{table}
\paragraph{Synaptic plasticity}
We adopt the STC model as synaptic plasticity model. Two conditions are needed for the occurrence of synaptic plasticity. First, the calcium concentration in the spine $[Ca^{2+}]_{s}$ should be high enough to set the spine into a plastic state, that is, synaptic tagging. Second, the plasticity related protein (PRP) is synthesized due to the calcium concentration in the dendrite $[Ca^{2+}]_{d}$. The calcium in the spine mainly comes from calcium-permeable synaptic receptors such as NMDA, while the calcium in the dendrite comes from the calcium ion channels on the cell membrane. We use ${z}_{l}$ and ${z}_{h}$ as the minimum and maximum factors that the synapse can achieve during LTD and LTP, respectively. The synapse change factor is determined by the following \eqref{Z}. We set ${z}_{l}$ = 0.5, ${z}_{h}$ = 5 in the experiments.
\begin{equation}
z=\frac{(1-{z}_{l})z_{h}e^{y}+z_{l}(z_{h}-1)e^{-y}}{(1-z_{l})e^y+(z_{h}-1)e^{-y}}
\label{Z}
\end{equation}
where the variable $y$ in \eqref{Z} is governed by the following \eqref{tau} to \eqref{alphap} :
\begin{equation}
\frac{dy}{dt}=\frac{Tag\cdot PRP}{{\tau}_{y}}
\label{tau}
\end{equation}
\begin{equation}
\frac{d(Tag)}{dt}=-{\alpha}_{T}Tag+{\beta}_{T}(Flag-Tag)
\end{equation}
\begin{equation}
Flag=
\begin{cases}
0& \text{$[Ca^{2+}]_{s}<Ca0_{s}$}\\ 
-1& \text{$Ca0_{s} \leq [Ca^{2+}]_{s} \leq Ca1_{s}$}\\
1& \text{$[Ca^{2+}]_{s}>Ca1_{s}$}\\
\end{cases}
\end{equation}
\begin{equation}
{\beta}_{T}=
\begin{cases}
0& \text{if Flag = 0}\\ 
{\beta}_{T,LTD}& \text{if Flag = -1}\\
{\beta}_{T,LTP}& \text{if Flag = 1}\\
\end{cases}
\label{betaT}
\end{equation}
Here, ${\tau}_{y}, {\alpha}_{T}$, ${\beta}_{T,LTD}$ and ${\beta}_{T,LTP}$ are the time constants. The variable $Ca0_{s}$ and $Ca1_{s}$ are calcium concentration thresholds of spine. We set ${\tau}_{y}$ = 1.0, ${\alpha}_{T}$ = 0.5, ${\beta}_{T,LTD}$ = 0.5, ${\beta}_{T,LTP}$ = 0.5, $Ca0_{s}$ = 0.1, $Ca1_{s}$ = 0.2 in the experiments. The PRP in Eq.(\ref{tau}) is defined as integration of ${PRP}_{rate}$, which satisfying the following \eqref{PRPRate}:
\begin{equation}
\begin{split}
\frac{d(PRP_{rate})}{dt}=&-\frac{PRP_{rate}}{\tau_{p}}+{\alpha}_{p}(1-{PRP}_{rate})\\
&-\frac{1}{4}(\frac{1}{{\tau}_{p}}+{\alpha}_{p})\int_{0}^{t}PRP_{rate}dt
\label{PRPRate}
\end{split}
\end{equation}
The value of ${\alpha}_{p}$ in \eqref{PRPRate} satisfies the following \eqref{alphap}:
\begin{equation}
{\alpha}_{p}=
\begin{cases}
0& \text{$[Ca^{2+}]_{d}<Ca0_{d}$}\\ 
{\alpha}_{T,LTD/LTP}& \text{$[Ca^{2+}]_{d} \geq Ca0_{d}$}\\
\end{cases}	
\label{alphap}
\end{equation}
Here, ${\alpha}_{T,LTD/LTP}$ is a time constant, ${\tau}_{p}$ is the decay time constant, $Ca0_{d}$ is calcium concentration threshold of dendrite. We set ${\alpha}_{T,LTD/LTP}$ = 0.000833, ${\tau}_{p}$ = 0.5, $Ca0_{d}$ = 0.12 in the experiments.
\subsection{Decoding module}
\label{LOCANNUpdate}
\begin{figure}[h]
	\centering
	\includegraphics[width=0.5\textwidth]{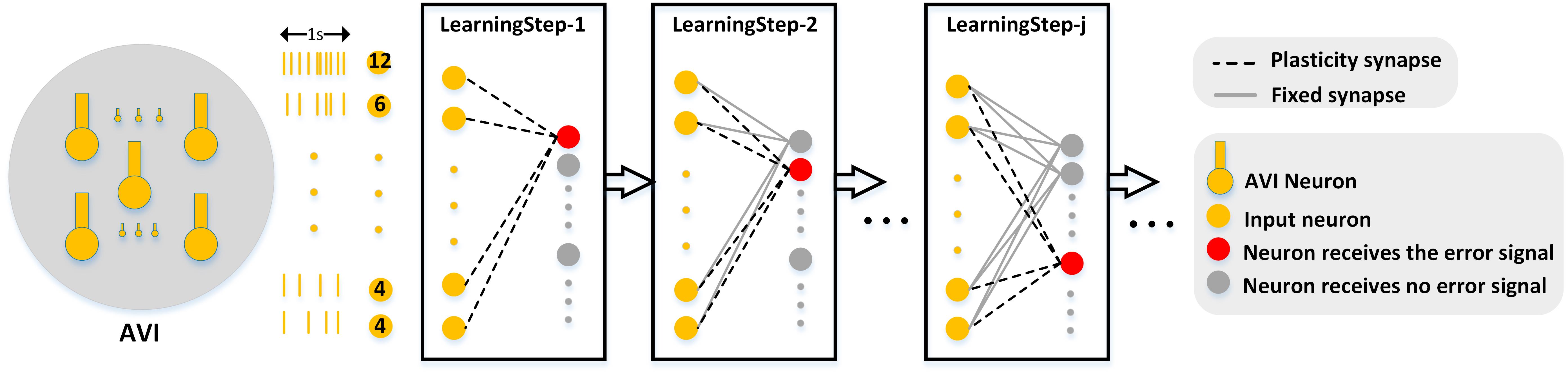}
	\caption{The procedure of the proposed updating algorithm for LOC-ANN during continual learning.  }
	\label{LOC-ANN}
	\vspace{-0.2cm}
\end{figure}
To decode the spike trains in AVI, we design an energy normalized linear output classifier based on the ANN called LOC-ANN for simplicity. Since LOC-ANN can suffer from catastrophic forgetting in the process of continual learning, we hence design the following algorithm to overcome this problem. 

We denote the dimension of input of LOC-ANN as $N_{av}$, the total number of objects or concepts need to be learned as $C$, the connection from the \emph{i-th} input neuron to the \emph{j-th} output neuron as ${W}(i,j)(i=1,2,...,N_{av}; j=1,2,...,C)$, the number of training samples in the \emph{k-th} class as $N_{k}$, and the AVI firing rate pattern of a sample that belongs to the \emph{k-th} class as $AVI_{j}^{(k)}$ ($j=1,2,...,N_{k}$), $AVI_{j}^{(k)}$$\in$$R^{1\times N_{av}}$.

At first, all the connections in LOC-ANN are set to 0. During learning the \emph{k-th} class, only the connections that belong to the \emph{k-th} output, denoted as ${W}(i,k)(i=1,2,...,N_{av})$, are able to update using the following \eqref{AverageFV}, while other connections keep fixed.
\begin{equation}
	{W}(i,k) = \frac{1}{N_{k}}\sum_{j=1}^{N_{k}}AVI_{j}^{(k)}(i)
	\label{AverageFV}
\end{equation}
After learning the \emph{k-th} class, we normalize connections that belong to the \emph{k-th} output neuron using \eqref{norm} to balance the energy of classification weights of each class:
\begin{equation}
	{W}_{norm}(i,k)= \frac{{W}(i,k)}{\sqrt{\sum_{j=1}^{N_{av}}W^{2}(j,k)}}
	\label{norm}
\end{equation}
Fig. \ref{LOC-ANN} shows the procedure of the proposed updating algorithm for LOC-ANN during learning. 
 
\section{Experimental design}
\label{ExperimentalDesign}
In this section, we will describe our work's experimental design, including the continual learning paradigm adopted in the experiments and the preprocessing of datasets.
\subsection{Continual learning paradigm}
\label{DefineCLParadigm}
In this paper, we focus on the incremental learning of new classes. Since the definition of incremental learning of new classes may be different in different papers, we will describe in detail the continual learning paradigm used in the present work based on \cite{hayes2018new}.

Supposed $D$ is the trainset. $C$ is the total number of classes in $D$. $D$ can be divided into $T$ individual batches that cannot be assumed to be iid, i.e., $D = \bigcup_{t=1}^{T}B_{t}$. Each batch $B_{t}$ consists of $N_{t}$ labeled training samples, i.e., $B_{t}=\{(x_{i},s_{i})\}_{i=1}^{N_{t}}$, where $x_{i}$ is a training sample and $s_{i}$ is the corresponding label. A commonly used continual learning paradigm satisfies that the model is only permitted to learn from batches sequentially in order, i.e., at time $t$ it only has access to $B_{t}$.

\begin{figure}[t]
	\centering
	\includegraphics[width=0.37\textwidth]{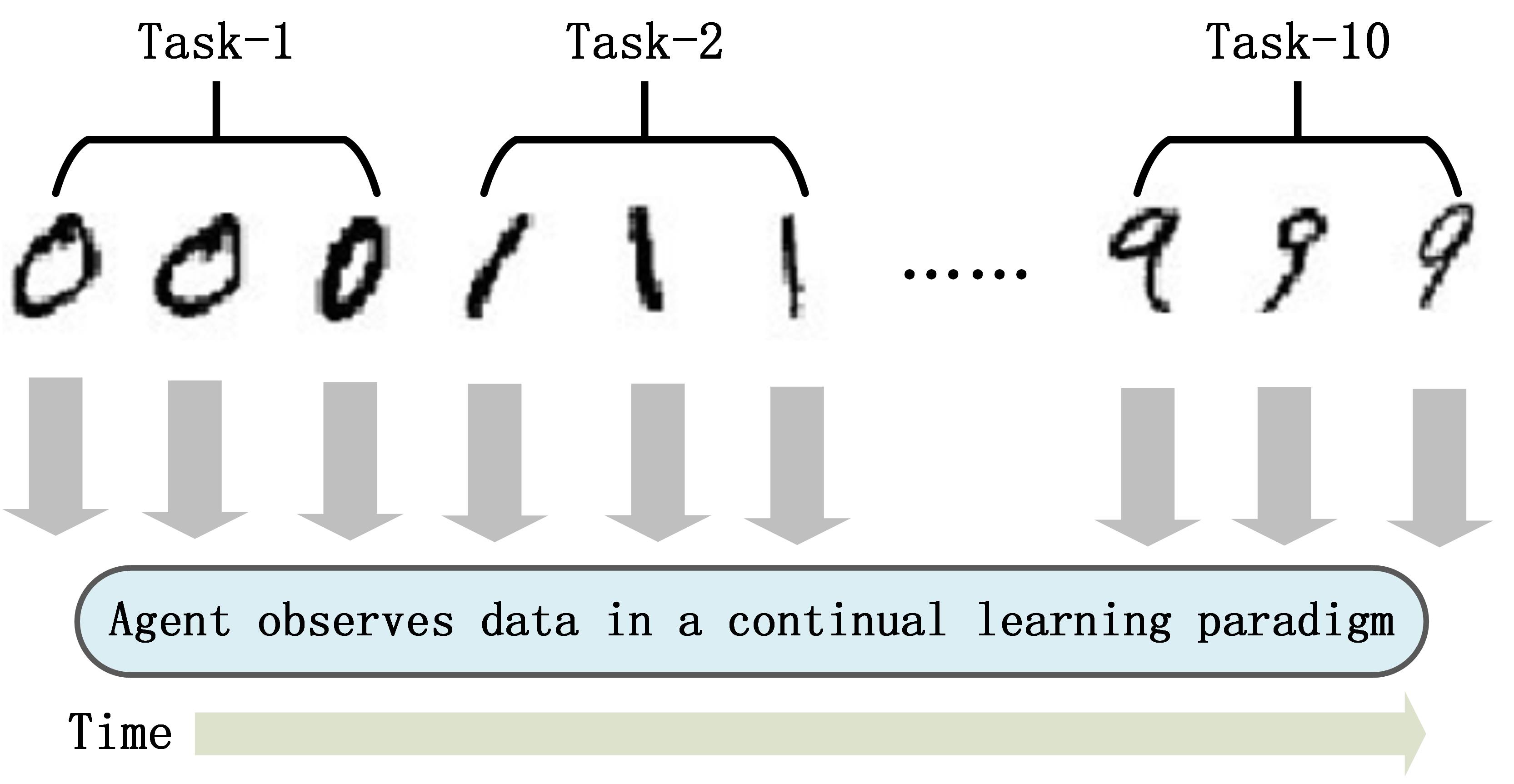}
	\vspace{-0.3cm}	
	\caption{ The continual learning paradigm adopted in the present work. }
	\label{CLParadigm}
\vspace{-0.3cm}	
\end{figure}

In the continual learning paradigm adopted in our work, we set $N_{t}$ = 1, which means that there is only one sample in each batch. Moreover, each sample can be observed only once, which is equal to setting the epochs number to 1. Under this continual learning paradigm, we present the samples in $D$ to AVIM in the order of their category indexes. See Fig. \ref{CLParadigm} for an example of such continual learning or incremental learning of new classes.

Since the continual learning paradigm described above may not be suitable for EWC\cite{kirkpatrick2017overcoming-r}, iCaRL \cite{rebuffi2017icarl:-m}, GEM \cite{lopezpaz2017gradient-m} and OWM \cite{zeng2018continuous-r}, which are state-of-the-art methods to be compared in this paper, we hence design the following continual learning paradigm that can be adopted in those methods: we set $T$ = $C$ and $B_{t}=\{(x_{i}^{(t)},t)\}_{i=1}^{N^{(t)}}$ ($t=1,2,...,C$), where $x_{i}^{(t)}$ is a training sample that belongs to the \emph{t-th} class and $N^{(t)}$ is the number of training samples in the \emph{t-th} class. For EWC, iCaRL, GEM and OWM, when learning a specific task, training samples from this task can be observed multiple times in order to achieve reasonable performance if needed.
\subsection{Datasets}
\paragraph{Constructing few-shot datasets}
We construct three few-shot image datasets of 10, 20 and 100 classes as experimental datasets from the following three frequently-used datasets: MNIST\cite{726791}, EMINST\cite{cohen2017emnist:} and CIFAR100\cite{Krizhevsky09learningmultiple}. Specifically, for MNIST10 and EMNIST20, we randomly select 50 samples per class from the trainset as training samples and 50 samples per class from the testset as test samples. For CIFAR100, we randomly select ten samples per class from the trainset as training samples and ten samples per class from the testset as test samples.

\begin{figure}[t]
	\centering
	\includegraphics[width=0.50\textwidth]{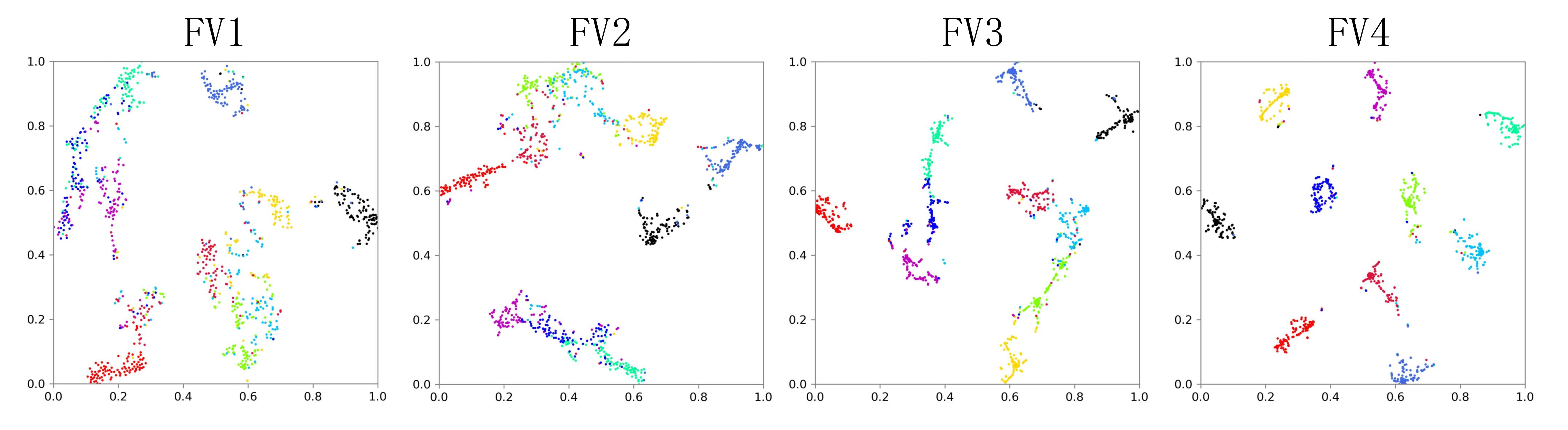}
	\caption{ T-SNE visualization of the four V-FV datasets of MNIST10. }
	\label{EMNIST20FV}
	\vspace{-0.4cm}
\end{figure}

\paragraph{Constructing the V-FV datasets} A V-FV dataset is a dataset composed of all the V-FVs that belong to the trainset and testset. For MNIST10 and EMNIST20, we use a LeNet-5-like \cite{lecun2015lenet} CNN to construct the corresponding V-FV datasets. For CIFAR100, we use a VGG-16 model \cite{simonyan2014very} to extract V-FVs of images. Since the topic we focus on is the multi-modal integration in MTL, we do not make any structural changes on these CNNs, except that we use the sigmoid activation function instead of ReLU in the last layer before the classification layer. Furthermore, for each dataset, we adopt CNNs with different qualities to obtain four V-FV groups with different levels of linear separability, which are denoted as FV1 to FV4. The linear separability levels of these V-FVs from FV1 to FV4 are increasing. See Fig \ref{EMNIST20FV} for T-SNE visualization \cite{van2008visualizing} of the V-FV datasets of MNIST10.
\section{Implementation details}
\label{ImplementationDetail}
 In this section, we will describe the implementation details of the proposed AVIM and other methods to be compared. All the experiments of AVIM are implemented in the in-house developed brain simulator named NiMiBrain, and all the ANN-based comparison experiments are implemented in PyTorch \cite{paszke2017automatic}.

\subsection{Details of AVIM}
\begin{figure}[t]
	\centering
	\includegraphics[width=0.48\textwidth]{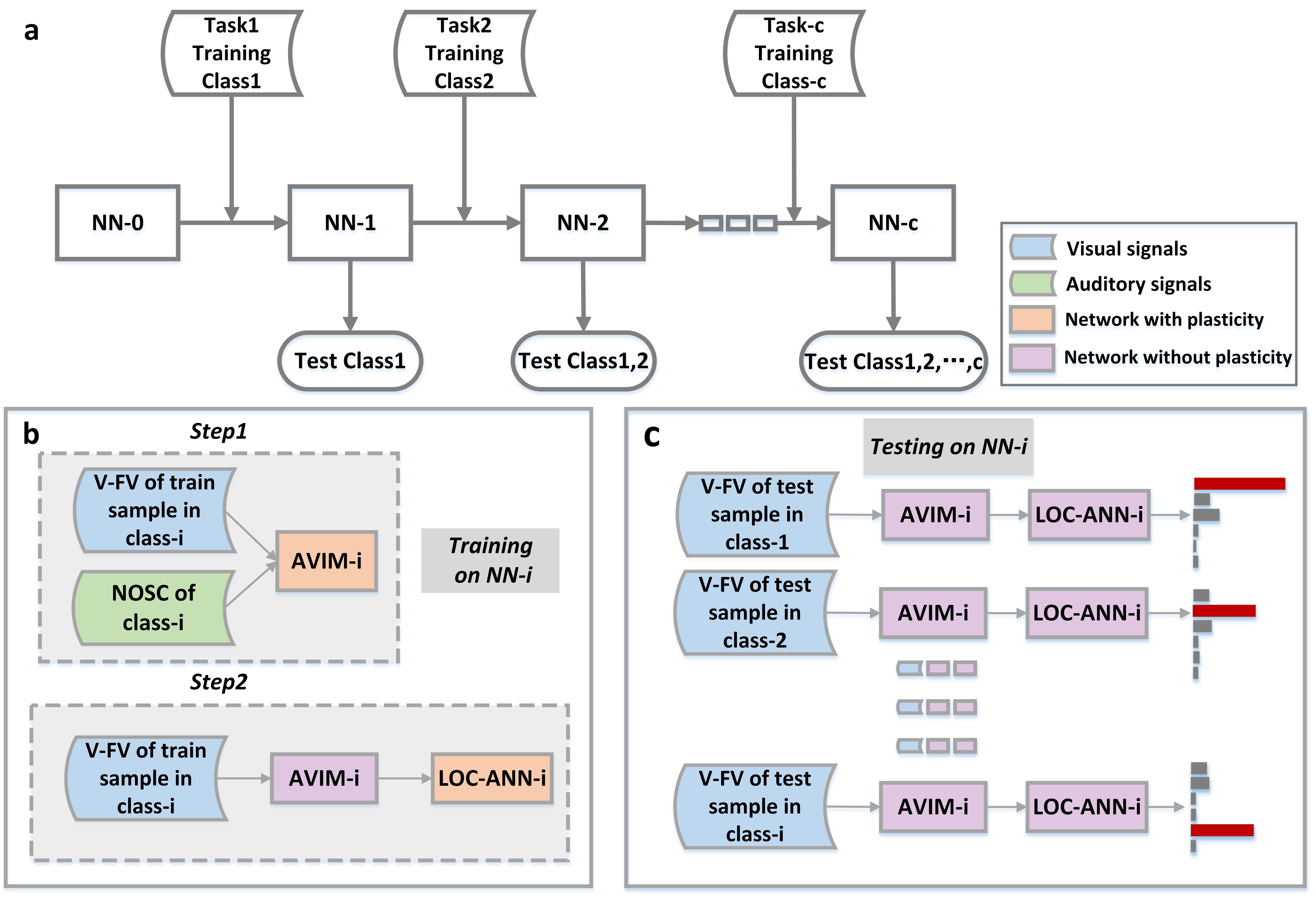}
	\vspace{-0.4cm}	
	\caption{\textbf{(a)}. The whole procedure of continual learning C classes. Here NN-i is composed of the AVIM-i and the LOC-ANN-i, and i represents the \emph{i-th} learning step. \textbf{(b)}. The detail of learning the \emph{i-th} class. \textbf{(c)}. The detail of testing after learning the \emph{i-th} class.}
	\label{experimentProcedure}
	\vspace{-0.5cm}	
\end{figure}

\paragraph{Size of AVIM} Since there are three datasets of different sizes, the sizes of AVIM adopted in different datasets are different; see TABLE \ref{experimentalSettings} for details.
\begin{table}[t]
	\caption{The sizes of AVIM in three different datasets.}
	\vspace{-0.2cm}
	\label{experimentalSettings}
	\centering
	\resizebox{85mm}{11mm}
	{
		\begin{tabular}{cccccccccc}
			\toprule
			\multicolumn{1}{c}{} &
			\multicolumn{1}{c}{\emph{CNN}} & 
			\multicolumn{4}{c}{\emph{NOSC}} &
			\multicolumn{4}{c}{\emph{AVIM}}	\\
			\cmidrule(r){2-10} 
			\emph{DATASET}     &\emph{$N_{vfv}$}	&\emph{$N_{nosc}$}	&\emph{C} &\emph{n}	&\emph{K} &\emph{$N_{v}$}	&\emph{$N_{a}$}	&\emph{$N_{av}$}	&\emph{$N_{i}$}\\
			\midrule
			MNIST10 &15	&15	&10 &3	&1   &15	&15	&50	&12 \\
			EMNIST20 &20 &20&20	&3	&1    &20	&20	&67	&16\\
			CIFAR100  &50&50&100	&5	&2  &50	&50	&167	&40\\
			\bottomrule
		\end{tabular}
\vspace{-1cm}	
}
\vspace{-0.5cm}	
\end{table}
\paragraph{Details of training}
For better description, we define a pair of NOSC and V-FV from the same class as an A-V training sample. When learning the \emph{i-th} class, each A-V training sample that belongs to this class is presented to AVIM for 2 seconds. Besides, there are 4 seconds between two adjacent training samples. When AVIM has finished learning the \emph{i-th} class, we keep it fixed. We only present the V-FVs that belong to the \emph{i-th} class in trainset to the model and then update the parameters in LOC-ANN using the algorithm described in Section \ref{LOCANNUpdate}. See a visualization of this process in Fig. \ref{experimentProcedure} (b).
\paragraph{Details of testing}
After learning the \emph{i-th} class, we keep AVIM and LOC-ANN fixed and only present the V-FVs of the learned classes in testset to the model for testing. Each test sample is presented for 1 second. And there are 0.1 seconds between presentations of two adjacent test samples. See Fig. \ref{experimentProcedure} (c) for a visualization of this process.

\subsection{Other methods to be compared}
We compare the proposed AVIM with six ANN methods, including ANN(Base), ANN(Offline), iCaRL\cite{rebuffi2017icarl:-m}, GEM \cite{lopezpaz2017gradient-m}, EWC \cite{kirkpatrick2017overcoming-r} and OWM \cite{zeng2018continuous-r}. Here, ANN(Base) represents the sequential training without any regularization and memory capacity, and ANN(Offline) represents the standard offline training that does not suffer from catastrophic forgetting. EWC and OWM are network regularization methods for continual learning. iCaRL and GEM are methods that based on memory-replay. For iCaRL, we compare the continual learning results of both iCaRL-NCM and iCaRL-CNN. Specifically, iCaRL-NCM adopts the nearest class mean (NCM) classifier for prediction, while iCaRL-CNN adopts the end-to-end classifier for prediction.

For a fair comparison, all ANN methods adopt a similar network structure as AVIM, which is a network with three fully-connected layers with $N_{v}$ inputs neurons, $N_{av}$ hidden neurons, and $C$ output neurons. We remove the biases in the classification layer. As for weight initialization, we use Kaiming-uniform \cite{he2015delving}. For ANN(Offline) and ANN(Base), we adopt abundant training on each task. For other ANN-based comparison methods, we use grid search for the training parameters such as learning rate and batch size and other parameters associated with particular methods such as the alpha in OWM. Fixed epochs number or early stop strategy are used in different comparison methods in order to get reasonable continual learning results. For memory-replay methods such as iCaRL and GEM, we set the memory capacity per class (MCPC) to 1. For other comparison methods except the ANN(Offline), the MCPC is 0. For ANN(Offline), all the training samples are used during learning. We run each comparison methods 10 times under different random seeds. 

\section{Results}
\label{Results}
\subsection{Numerical results}
The continual learning performance of each comparison method is evaluated by the final top-1 test accuracy of all the learned classes. Statistic data of these comparison results are recorded in TABLE \ref{CLResultTable} to TABLE \ref{CLResultTable_2}. In addition, we visualize these methods' test accuracy curves during continual learning on the few-shot CIFAR100-FV datasets, see Fig. \ref{AccuracyCurve} for details. 
\begin{table}[h]
\centering
\vspace{-0.3cm}
\caption{Final continual learning accuracies of AVIM and other ANN methods on MNIST-FV.  
\textbf{Black bolding} in the table represents the best continuous learning performance in each experiment under the same constrain of memory capacity per class.}	
\vspace{-0.3cm}
\resizebox{90mm}{19mm}
{
	\begin{tabular}{cccccc}
		\toprule
		Method & MCPC &FV1($\%$)&   FV2($\%$)&  FV3($\%$) &FV4($\%$)\\
		\midrule
		ANN(Offline)&All &71.8$\pm$1.0 &75.0$\pm$0.9 &85.4$\pm$0.6 &88.9$\pm$0.5	\\
		\midrule
	    ANN(Base)   &0  &10.1$\pm$0.1 &10.4$\pm$0.9 &10.6$\pm$1.5 &12.3$\pm$3.1    \\
	    EWC &0         &32.2$\pm$5.5 &34.4$\pm$6.6 &43.6$\pm$9.2 &52.5$\pm$10.3   \\
	    OWM &0 &62.5$\pm$0.5 &68.5$\pm$1.0 &\textbf{83.9$\pm$0.9} &88.8$\pm$0.6 \\
	    AVIM &0 &\textbf{63.3$\pm$1.9} &\textbf{70.5$\pm$0.8} &82.5$\pm$2.0 &\textbf{89.8$\pm$0.2} \\
	    \midrule
	    GEM &1   	 &56.5$\pm$3.6 &66.8$\pm$3.2 &79.5$\pm$2.7 &89.5$\pm$0.3	\\
	    iCaRL-CNN &1   &54.5$\pm$1.2 &62.8$\pm$0.6 &69.9$\pm$1.2 &88.4$\pm$0.5	\\
	    iCaRL-NCM &1	 &\textbf{63.8$\pm$0.6} &\textbf{71.6$\pm$0.8} &\textbf{83.2$\pm$0.4} &\textbf{89.8$\pm$0.3}\\	   
		\bottomrule
		\label{CLResultTable}
		\vspace{-0.4cm}
	\end{tabular}
}
\end{table}
\begin{table}[!htb]
	\vspace{-0.3cm}	
	\centering
	\caption{Final continual learning accuracies of AVIM and other ANN methods on EMNIST20-FV.}
	\vspace{-0.3cm}	
	\resizebox{90mm}{19mm}
	{
		\begin{tabular}{cccccc}
			\toprule			
			Method &MCPC &FV1($\%$) &FV2($\%$) & FV3($\%$) &FV4($\%$)\\
			\midrule
			ANN(Offline)&All &72.8$\pm$0.8 &77.5$\pm$0.5 &83.3$\pm$0.5 &90.0$\pm$0.4	\\
			\midrule
			 ANN(Base)   &0  &9.7$\pm$0.1 &9.7$\pm$0.1 &9.6$\pm$0.1 &10.0$\pm$0.5    \\
			EWC       &0   &26.3$\pm$3.8 &27.4$\pm$3.6 &29.9$\pm$3.2 &32.7$\pm$3.1   \\
			OWM &0 &62.7$\pm$0.6 &\textbf{69.5$\pm$0.9} &\textbf{79.9$\pm$0.5} &90.0$\pm$0.4 \\
			AVIM	&0	 &\textbf{63.6$\pm$0.3} &67.5$\pm$1.1 &76.6$\pm$0.5 &\textbf{90.8$\pm$0.5}	\\	
			\midrule
			GEM     &1	 &52.6$\pm$3.6 &60.1$\pm$3.0 &72.9$\pm$1.6 &89.0$\pm$0.4	\\
			iCaRL-CNN &1   &57.8$\pm$0.6 &63.3$\pm$0.6 &73.6$\pm$0.6 &89.1$\pm$0.4	\\
			iCaRL-NCM &1	 &\textbf{63.4$\pm$0.6} &\textbf{71.4$\pm$0.5} &\textbf{79.5$\pm$0.4} &\textbf{89.8$\pm$0.3}	\\							
			\bottomrule
			\label{CLResultTable_1}
		\vspace{-0.4cm}
		\end{tabular}
	\vspace{-0.5cm}
	}
\end{table}
\begin{table}[!htb]
	\vspace{-0.3cm}
	\centering
	\caption{Final continual learning accuracies of AVIM and other ANN methods on CIFAR100-FV.}	
	\vspace{-0.3cm}
	\resizebox{90mm}{19mm}
	{
		\begin{tabular}{cccccc}
			\toprule			
			Method &MCPC &FV1($\%$) &FV2($\%$) &FV3($\%$) &FV4($\%$)\\
			\midrule
			ANN(Offline)&All &49.8$\pm$0.3 &57.5$\pm$0.4 &66.1$\pm$0.3 &72.9$\pm$0.2	\\
			\midrule
			 ANN(Base)   &0  &1.9$\pm$0.5 &2.9$\pm$0.6 &3.3$\pm$0.3 &3.0$\pm$0.3    \\
			EWC       &0    &13.6$\pm$2.3 &16.4$\pm$2.4 &18.8$\pm$2.2 &27.6$\pm$3.7	\\
			OWM &0 &38.9$\pm$0.7 &47.5$\pm$0.9 &58.0$\pm$1.2 &71.4$\pm$0.5 \\
			AVIM	&0	  &\textbf{47.9$\pm$0.2} &\textbf{56.4$\pm$0.8} &\textbf{64.8$\pm$0.6} &\textbf{72.9$\pm$0.3}	\\
			\midrule
			GEM     &1	  &36.5$\pm$1.8 &45.2$\pm$1.7 &55.2$\pm$1.3 &72.0$\pm$0.3	\\
			iCaRL-CNN &1    &44.5$\pm$0.7 &52.7$\pm$0.5 &61.7$\pm$0.7 &71.9$\pm$0.4	\\
			iCaRL-NCM &1	  &\textbf{47.0$\pm$0.7} &\textbf{55.4$\pm$0.6} &\textbf{63.9$\pm$0.4} &\textbf{72.9$\pm$0.2}	\\						
			\bottomrule
			\label{CLResultTable_2}
		\end{tabular}
	\vspace{-0.9cm}
	}
\vspace{-0.8cm}
\end{table}

\subsection{Result analysis}
\begin{figure}[t]
	\centering
	\includegraphics[width=0.52\textwidth]{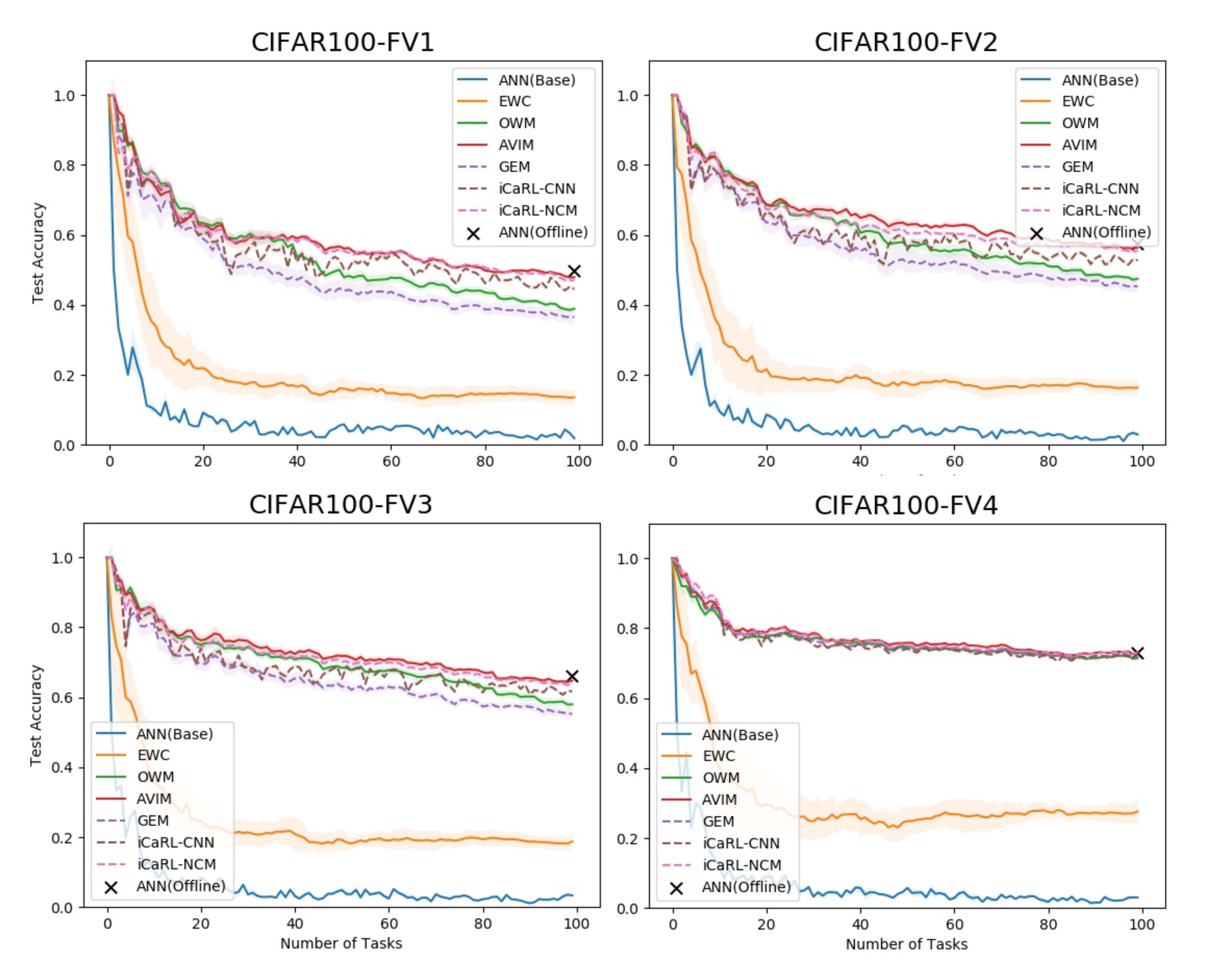}
	\caption{The continual learning test accuracy curves of AVIM and other ANN methods on the few-shot CIFAR100-FV datasets. The solid lines in this figure represent methods that do not need to store memory, and the dotted lines represent methods that need to store memory for replaying during learning. }
	\label{AccuracyCurve}
	\vspace{-0.5cm}
\end{figure}

From the results shown in TABLE \ref{CLResultTable}-\ref{CLResultTable_2} and Fig. \ref{AccuracyCurve}, we can conclude that: 

1. In general, the proposed AVIM can achieve comparable performance as OWM and iCaRL, while being much better than GEM and EWC. 

2. On the few-shot CIFAR100-FV datasets, the performance of AVIM can surpass that of OWM especially in the FV1 to FV3 datasets. The reason of this phenomenon could be that the weight space (50$\times$167+167$\times$100) is too limited for OWM to find enough orthogonal sub-spaces for learning 100 classes. Under this limited weight space, OWM could not update the weights in a way without seriously interference with previous learned classes when learning on new classes, especially in the case that FVs of different classes are non-linearly separable.

3. On the few-shot CIFAR100-FV datasets, it seems that the performance of iCaRL-NCM is slightly worse than that of AVIM, the reason of which is that the limited memory capacity of iCaRL affects its performance when learning more classes. Additional experiments have shown that for iCaRL, the more memory capacity it has, the better continual learning performance it can achieve. 

4. For iCaRL, iCaRL-NCM can achieve better performance compared with iCaRL-CNN, which means that NCM can serve as a more efficient classifier than the end-to-end classifier during this continual learning paradigm. It is worth mentioning that the NCM classifier adopted in iCaRL-NCM normalizes the mean vector of each class, which is similar to the equal-energy assumption of each class we made in the proposed LOC-ANN. 

5. GEM, another advanced method that based on memory-replay, is overall worse than iCaRL in the comparison results. The reason of these results is largely because GEM randomly selects samples from the previously learned classes to replay, while iCaRL adopts a well-designed sample selection method. Moreover, like iCaRL, the limited memory capacity of GEM can also affect its performance on memory consolidation. In short, memory quality and capacity can have significant impacts on the continual learning performance of memory-replay based methods such as iCaRL and GEM, especially when learning more classes. 

6. The performance of EWC is significantly worse than AVIM, OWM, iCaRL, and GEM, suggesting that using EWC alone for protecting important weights is not sufficient to overcome catastrophic forgetting when incremental learning of new classes. 

\begin{figure}[t]
	\centering
	\includegraphics[width=0.5\textwidth]{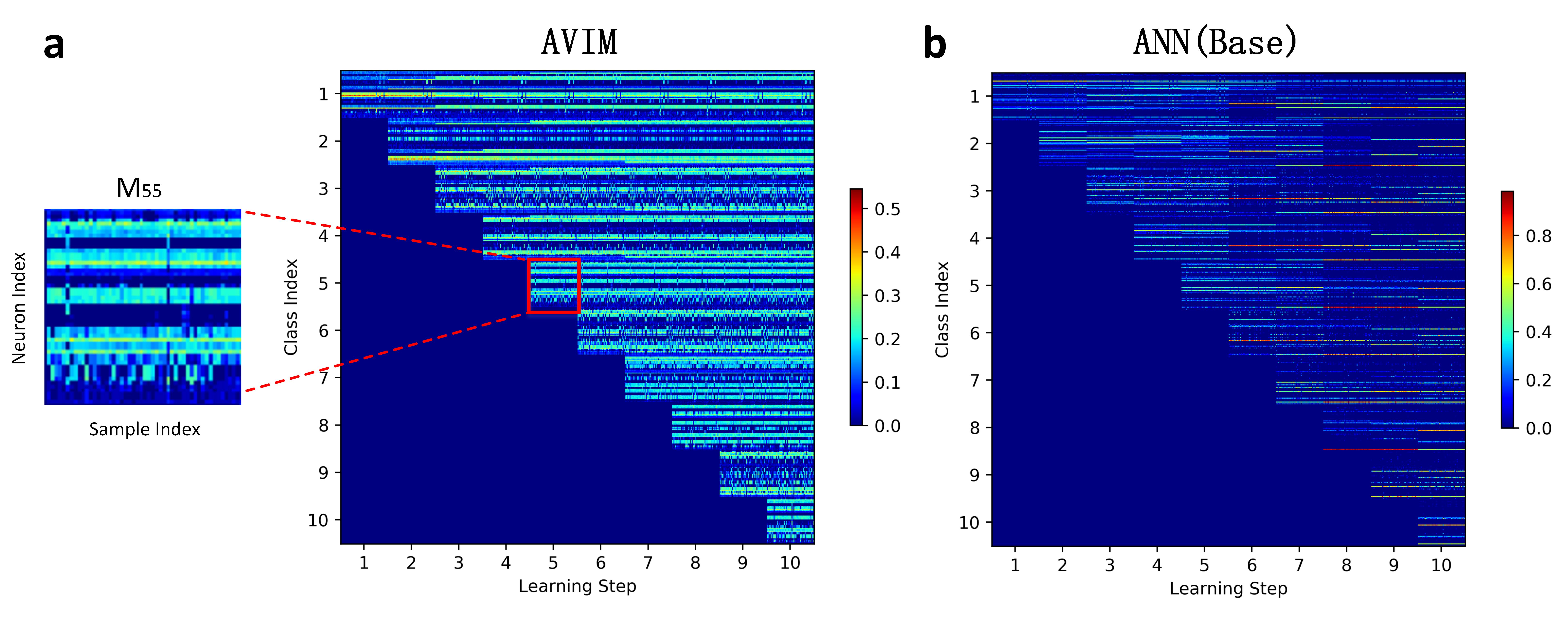}
	\vspace{-0.5cm}
	\caption{In the MNIST10-FV1 dataset, we visualize \textbf{(a)}. AVI firing rate patterns of samples in each learned class during continual learning. The small matrix ${M}_{i,j}$ ($1\le i \le 10$; $i\le j\le 10$) on the left is composed of the AVI firing rate patterns belong to the \emph{i-th} class in the \emph{j-th} learning step. The large matrix on the right is composed of all the ${M}_{i,j}$ during continual learning. \textbf{(b)}. ANN(Base)'s hidden patterns of samples in each learned class during continual learning. All these patterns are normalized to vectors whose magnitude are 1 for visualization.}
	\label{AVIMPattern}
\vspace{-0.6cm}
\end{figure} 
\section{Discussion}
\label{Discussion}
In this section, we discuss the advantages and possible limitations of the three important components in the proposed model.
\paragraph{AVIM}
AVIM is a spiking neural network, served as the core component for audio-visual integration and concept formation. Experiments (see Fig. \ref{AVIMPattern}) show that AVIM can generate stable AVI representations of objects during continual learning, while ANN(Base)'s hidden patterns are relatively instable, indicating that BP without extra restrictions could be too plastic to consolidate memory. The success of formation of these stable representations of objects is mainly due to the multi-compartmental neuron model and the STC plasticity. Additional experimental results show that if the multi-compartment neuron models in AVIM are replaced with the point neuron model, AVI could not form stable representations of objects, which results in catastrophic forgetting of previous learned classes under the same framework. Although two-compartments neuron model is used in our experiments, more compartments and connection patterns should be considered in the future. Besides, we use STC plasticity, which is a calcium-based local learning rule, allowing us to study and control more details of neuronal responses. It will also be worthwhile to try different SNN learning rules to see what would happen under this experimental setting.
\paragraph{NOSC} In this paper, we define NOSC as the high-level auditory feature codes, and use them as the supervised signals to guide the audio-visual integration. Besides, we do not allow the connections from AF, which take NOSC as input, to AVI to have plasticity during learning. Although this design is easy to implement and is reasonable to some extent, it is more biologically plausible to integrate multi-modal signals in the case of both visual and auditory plasticity.
\paragraph{LOC-ANN}
LOC-ANN is used to decode the spike trains of AVI and get the final classification results. We design LOC-ANN to be an energy normalized linear classifier based on the assumption that different classes during learning have equal energy. Furthermore, we make LOC-ANN update in a flexible way in the continual learning paradigm adopted in this paper. Overall, LOC-ANN is an efficient decoder for AVIM to obtain the classification results. 
\section{Conclusion}
\label{Conclusion}
In this paper, we propose a novel biologically plausible audio-visual integration model (AVIM) with multi-compartment neurons and STC plasticity for continual learning. Our experimental results show that AVIM can achieve comparable performance with advanced methods such as OWM, iCaRL and GEM in different datasets. Moreover, it can generate stable representations of objects during continual learning. It should be noted that the present work does not mainly aim to get the top-1 performance on specific datasets but more to explore the possible mechanism of brain-like learning and concept formation. Finally, we point out that the proposed AVIM is just an initial step towards solving the problem of concept formation and continual learning. There is still a long way to go.
\section*{Acknowledgements}
This work is supported by the the Fundamental Research Funds for the Central Universities (grant no. CUC18A001) and the Open Project Program of the State Key Laboratory of Mathematical Engineering and Advanced Computing (grant no. 2020A09).
\bibliographystyle{IEEEtran}
\bibliography{citation}
\end{document}